\documentclass[runningheads]{llncs}
\usepackage[T1]{fontenc}
\usepackage{graphicx}
\usepackage{amsmath}
\usepackage{tikz}
\usepackage{booktabs}
\usepackage{multirow} 
\usepackage{siunitx}  
\usepackage{placeins}
\usepackage{framed}
\usepackage{float}
\usepackage{amsfonts}
\usepackage{amssymb}
\usepackage{physics}
\usetikzlibrary{shapes.geometric, arrows.meta, positioning, fit, backgrounds, calc}

\begin{document}

\title{Quantum Machine Learning for Colorectal Cancer Data: Anastomotic Leak Classification and Risk Factors}

\author{Vojt\v{e}ch Nov\'{a}k\inst{1,2} \and
Ivan Zelinka\inst{1,2} \and
Lenka P\v{r}ibylov\'{a}\inst{3} \and
Lubom\'{i}r Mart\'{i}nek\inst{4,5} \and
Vladim\'{i}r Ben\v{c}urik\inst{5} \and 
Martin Beseda\inst{6}}

\authorrunning{V. Nov\'{a}k et al.}

\institute{Department of Computer Science, Technical University of Ostrava, Czech Republic \and
IT4Innovations National Supercomputing Center, Ostrava, Czech Republic \and
Department of Applied Mathematics, Technical University of Ostrava, Czech Republic \and
Department of Surgical Studies, Faculty of Medicine, University of Ostrava, Czech Republic \and
Department of Surgery, University Hospital Ostrava, Czech Republic \and 
Dipartimento di Ingegneria e Scienze dell'Informazione e Matematica, Universit\`{a} dell'Aquila, Via Vetoio, I-67010 Coppito, L'Aquila, Italy}

\maketitle              

\begin{abstract}
This study evaluates colorectal risk factors and compares classical models against Quantum Neural Networks (QNNs) for anastomotic leak prediction. Analyzing clinical data with 14\% leak prevalence, we tested ZZFeatureMap encodings with RealAmplitudes and EfficientSU2 ansatze under simulated noise. $F_\beta$-optimized quantum configurations yielded significantly higher sensitivity (83.3\%) than classical baselines (66.7\%). This demonstrates that quantum feature spaces better prioritize minority class identification, which is critical for low-prevalence clinical risk prediction. Our work explores various optimizers under noisy conditions, highlighting key trade-offs and future directions for hardware deployment.

\end{abstract}

\section{Introduction}

Colorectal cancer (CRC) is a leading cause of cancer-related mortality globally, with notably high incidence in the Czech Republic \cite{majkur2025cancer}. Approximately 8,000 new cases are diagnosed annually in the region, resulting in nearly 4,000 deaths \cite{dusek2020crc}. Surgical resection is the primary curative intervention for CRC; however, it carries a substantial risk of severe postoperative complications \cite{ruyk2021risk}. Anastomotic leak (AL) is one of the most critical mechanical failures in this domain, occurring when surgically reconnected bowel tissue fails to heal adequately \cite{isgrc2010def}. This structural breach allows intestinal contents to leak into the abdominal cavity, frequently resulting in life-threatening peritonitis and sepsis \cite{bruce2001systematic}.

The etiology of AL is multifactorial, driven by both patient-specific comorbidities and intraoperative surgical mechanics \cite{kingham2008colonic}. Systemic conditions such as Diabetes Mellitus (DM) and a history of smoking directly impair microvascular health, tissue oxygenation, and subsequent wound healing. To mitigate these risks, surgeons employ precise intraoperative interventions, such as the preservation of the left colic artery (ACSP) to maintain adequate blood perfusion to the anastomotic site, and the insertion of a transanal drain (NoCoil) to mechanically decompress intraluminal pressure. Furthermore, recent advancements in intraoperative imaging utilize Indocyanine Green (ICG) fluorescence angiography. ICG is a water-soluble fluorescent dye administered intravenously; under near-infrared illumination, it enables real-time visualization of vascular patterns, allowing surgeons to dynamically assess tissue perfusion and optimize the anastomotic line. A recent study by Ben\v{c}ur\'{i}k et al. \cite{benvcurik2021intraoperative} demonstrated that ICG application significantly reduces AL rates in rectal cancer surgeries.

Predicting AL risk presents a significant classification challenge due to class imbalance, with AL occurring in approximately $14\%$ of our clinical cohort. This skew necessitates models that maintain high sensitivity for life-threatening false negatives without excessively inflating false-positive rates. Consequently, feature selection in our model is restricted to clinically and physiologically interpretable variables rather than purely statistical abstractions.

This dataset complexity motivates the exploration of non-classical hypothesis spaces. While classical Machine Learning (ML) serves as a foundation for healthcare analytics \cite{carleo2018constructing,liu2019unitary}, Quantum Computing (QC) offers an alternative framework for complex pattern recognition \cite{biamonte2017quantum}. Quantum Machine Learning (QML) \cite{ciliberto2018quantum} leverages the principles of quantum mechanics, mapping classical data into high-dimensional Hilbert spaces to identify correlations that may elude classical architectures \cite{schuld2019quantum}. The theoretical advantage of deploying parameterized quantum circuits over classical architectures is grounded in computational complexity. If a data encoding map lacks entangling operations, the resulting quantum state is strictly separable and the inner products can be efficiently simulated by classical algorithms. By employing entangling operations within the feature map, the circuit is capable of exploring the full $2^n$ dimensional Hilbert space. When the circuit depth and qubit connectivity reach a sufficient level of complexity, estimating the fidelity of these states becomes classically intractable and is formally classified as a \#P hard problem \cite{havlicek2019supervised}. This strict complexity theoretic bound guarantees that the quantum model operates within a mathematically distinct hypothesis space rather than acting as a dequantizable surrogate.

In the current Noisy Intermediate-Scale Quantum (NISQ) era, deep quantum circuits are heavily restricted by hardware decoherence and gate errors \cite{lau2022nisq,preskill2018quantum}. Consequently, modern QML relies on hybrid approaches such as the Variational Quantum Classifier (VQC) \cite{havlicek2019supervised}. These Variational Quantum Algorithms utilize parameterized quantum circuits, or Quantum Neural Networks (QNNs), which are optimized iteratively via classical loops \cite{cerezo2021variational,abbas2021power}. QNNs have shown early promise in supervised learning tasks across various biomedical domains, from omics and gene classification to medical imaging \cite{li2021quantum,sengupta2021quantum,maheshwari2023quantum}. However, their empirical advantage over classical models on noisy, imbalanced clinical datasets is not yet established \cite{liu2021rigorous,schreiber2023classical}.

Beyond biomedical applications, variational quantum algorithms are utilized in remote sensing \cite{gupta2022how}, topological state identification \cite{ciaramelletti2025detecting}, combinatorial optimization \cite{trovato2025preliminary}, and quantum chemistry \cite{beseda2024state,illesova2025transformation}. Physical deployment of these algorithms requires rigorous hardware characterization, including unitary channel discrimination \cite{bilek2026experimental} and measurement certification \cite{lewandowska2025benchmarking}. A remaining open problem is evaluating the robustness of these algorithms when applied to highly imbalanced classical datasets under realistic noise models.

In this paper, we performed benchmarking of simulated QNNs against hyperparameter-tuned classical models for AL prediction. Evaluating model performance under simulated quantum noise, we systematically compare data encoding maps, variational ansatze, and optimization routines. Our analysis quantifies the trade-off between threshold-dependent classification metrics and probability calibration. Additionally, we apply perturbation-based feature importance to compare the learned representations of QNNs and classical linear models, establishing baseline requirements for future physical hardware deployment.

\section{Methodology}

We first introduce the medical problem in detail, define the clinical cohort, and outline all patient variables, including surgical techniques, patient history, and physiological markers with their respective units. Prior to any machine learning tasks, we perform rigorous statistical analysis and feature reduction to isolate the most significant clinical predictors. Building on this clinical foundation, we introduce the quantum machine learning methodology, detailing the design of our parameterized quantum circuits using specific data encoding maps and variational ansatze that we compare with classical machine learning models.

\subsection{Clinical Dataset and Feature Space}
Our study utilizes data collected from the Surgical Department of Hospital Nový Jičín a.s. between 2015 and 2016, comprising 200 patients (28 with an anastomotic leak, 172 without). The primary goal of this study was to identify the statistical significance of intraoperative techniques such as NoCoil, ACSP, PERFB, and ICG, and to identify potential risk factors for anastomotic leak occurrence. Initial exploratory analysis categorized the explanatory variables into patient medical history (comorbidities) and intraoperative surgical techniques.

The statistical significance of each predictor was assessed using a Chi-squared ($\chi^2$) goodness-of-fit test. For patient comorbidities, Diabetes Mellitus (DM) and smoking were identified as the most critical risk factors. The presence of DM increases the relative risk of leak occurrence by 2.16 times ($p=0.036$), with leaks occurring in 9 out of 36 patients. Similarly, smoking increases the risk by 2.31 times ($p=0.042$), affecting 9 out of 34 patients. In patients presenting with either of these factors, the incidence of anastomotic leak approaches 25\%.

Regarding surgical interventions, the insertion of a transanal drain (NoCoil) provides a statistically significant protective effect, reducing leak occurrence by a factor of 3.16 ($p=0.032$), as leaks occurred in only 3 out of 55 patients when the drain was used compared to 25 out of 145 when it was not. Intraoperative fluorescence imaging (ICG) also demonstrated a significant univariate association, reducing leaks by 2.11 times ($p=0.042$), with 9 out of 100 patients experiencing a leak in the ICG group versus 19 out of 100 in the control. While the preservation of the left colic artery (ACSP) showed a marginal protective trend individually ($p=0.074$), with 5 leaks out of 65 patients, it was prioritized during multivariate feature selection.

Because NoCoil and ICG are frequently applied concurrently in clinical practice when a leak is suspected, they exhibit overlapping effects that dilute ICG's independent predictive value in a multivariate setting. Step-wise feature reduction utilizing Akaike's Information Criterion (AIC) across multivariate logistic regression models resulted in the exclusion of ICG to optimize the AIC score. Consequently, to mitigate overfitting and focus on the computational properties of the quantum models, the feature space was strictly bounded to four predictors: DM, Smoking, NoCoil, and ACSP, which provided the most stable, independent contributions to the model's variance.

\subsection{Quantum Circuit Architecture}

To process classical clinical data within a quantum framework, we employed a 4-qubit parameterized quantum circuit consisting of a quantum feature map and a variational ansatz. The fundamental mechanism of quantum machine learning relies on mapping classical data into a high-dimensional complex Hilbert space. Let the classical data vector be denoted as $x \in \mathbb{R}^d$, where $d$ represents the four clinical features. This vector is embedded into the Hilbert space $\mathcal{H}$ by applying a parameterized unitary operator $U_{\Phi}(x)$ to an initial ground state $\ket{0}^{\otimes n}$ to produce the quantum state
\begin{equation}
\ket{\Phi(x)} = U_{\Phi}(x)\ket{0}^{\otimes n},
\end{equation}
where $n=4$ is the number of qubits.We achieved this mapping using the ZZFeatureMap, which is a second-order expansion defined by the unitary

\begin{equation}
U_{\Phi(x)} = \exp\left(i \sum_{j,k \in \mathcal{E}} \phi_{j,k}(x) Z_j Z_k\right) \exp\left(i \sum_{j=1}^n \phi_j(x) Z_j\right) H^{\otimes n},
\end{equation}
where $H$ is the Hadamard gate for superposition, $Z$ is the Pauli-Z operator, $\phi_j(x) = x_j$ represents individual features, and $\phi_{j,k}(x) = (\pi - x_j)(\pi - x_k)$ captures interactions between features $j$ and $k$ using CNOT gates to create entanglement.

Following the feature map, we applied a variational ansatz $V(\theta)$ to perform the classification. We compared two architectures. The first is RealAmplitudes (RA), which uses layers of $R_y(\theta)$ gates that rotate the qubits around the y-axis by trainable parameters $\theta$ and connects them with CNOT gates to keep the state amplitudes in the real domain. The second is EfficientSU2 (ESU2), which increases model expressivity by applying both $R_y(\theta)$ and $R_z(\theta)$ rotations followed by CNOT entanglement layers. In both cases, the final state is given by

\begin{equation}
\ket{\psi(x, \theta)} = V(\theta)\ket{\Phi(x)},
\end{equation}
where $\theta$ represents the vector of weights optimized during training to minimize the classification error.

This enables the quantum circuit to capture complex and non-linear feature interactions that mimic polynomial kernels. The inner product between two such mapped quantum states defines a quantum kernel function \cite{havlicek2019supervised} evaluated on two classical data points $x$ and $y \in \mathbb{R}^d$, which is formally expressed as
\begin{equation}
K(x,y) = |\braket{\Phi(x)}{\Phi(y)}|^2,
\end{equation}
where $K(x,y)$ quantifies the similarity between the data points in the expanded feature space \cite{havlicek2019supervised}. For highly imbalanced medical datasets, this non-linear mapping into a higher-dimensional Hilbert space is critical. By utilizing the quantum kernel trick, the feature map projects the clinical data into a high-dimensional state space where the minority class (anastomotic leak occurrences) can become more linearly separable from the majority class. This transformation allows the variational circuit to define a hyperplane in Hilbert space that corresponds to a complex, non-linear decision boundary in the original feature space, which is essential for capturing the rare conditions leading to a leak \cite{schuld2019quantum}. The fundamental properties of these encoding circuits, specifically their expressivity and entangling capacity, heavily influence the model's ability to resolve these minority class boundaries across different feature dimensions \cite{illesova2025qmetric}.

Following state preparation, the encoded quantum information is processed through trainable parameterized layers. Let $W(\theta)$ denote the unitary operation of the variational ansatz, where $\theta \in \mathbb{R}^m$ is the vector of $m$ trainable weights, representing the total number of rotation angles in the circuit, optimized during the classical training loop. The final quantum state prior to measurement is given by
\begin{equation}
\ket{\psi(x, \theta)} = W(\theta)U_{\Phi}(x)\ket{0}^{\otimes n},
\end{equation}
where $U_{\Phi}(x)$ remains the data encoding unitary. The classification output is derived from the expectation value of an operator $\hat{M}$ acting on the Hilbert space. In our implementation, $\hat{M}$ is the Pauli-Z operator $Z = \ket{0}\bra{0} - \ket{1}\bra{1}$ applied to the first qubit, which serves as the designated readout qubit. The continuous output of the model is therefore defined as
\begin{equation}
f(x, \theta) = \bra{\psi(x, \theta)} \hat{M} \ket{\psi(x, \theta)},
\end{equation}
where $f(x, \theta) \in [-1, 1]$ represents the expected value of the measurement. Because this raw output falls between $-1$ (class 1) and $1$ (class 0), it must be transformed into a probability. We use a classical sigmoid link function, $\sigma(z) = (1 + e^{-z})^{-1}$, to map these continuous values to the interval $[0, 1]$. This allows the model to predict the probability of an anastomotic leak and enables the use of standard binary cross-entropy as the loss function during optimization.

The choice of ansatz dictates the balance between model expressibility and trainability. We benchmarked two distinct architectures to evaluate this computational overhead. The RealAmplitudes ansatz consists of alternating layers of $R_y$ rotations gates and CNOT entangling gates, restricting the parameter space to real numbers and yielding a shallow circuit depth of 11 at 3 repetitions. The EfficientSU2 ansatz is a more expressive architecture utilizing single-qubit rotations around multiple axes ($R_y$ and $R_z$) prior to entanglement, which captures more intricate quantum states but increases circuit to a depth of 15 and imposes a heavier optimization burden due to the larger parameter vector $\theta$.

\subsection{Noise Simulation and Optimization}
To replicate the operational realities of NISQ hardware, all quantum circuits were simulated using Qiskit's AerSimulator \cite{qiskit2024} equipped with depolarizing noise model. We applied a conservative single-qubit gate error probability of $p_{gate} = 0.05$, distributed evenly across Pauli X, Y, and Z operations, alongside sampling noise simulated at 1024 shots per circuit evaluation.

Variational parameters were updated iteratively utilizing both gradient-based (BFGS \cite{liu1989limited}, SLSQP \cite{stoer1985principles}) and gradient-free metaheuristic (CMA-ES \cite{hansen2001completely}, COBYLA \cite{powell2007view}, SPSA \cite{spall1992multivariate}) optimizers. To ensure statistical reliability against the stochastic measurement processes, all performance metrics were averaged across 10 independent optimization runs for each ansatz-optimizer combination.

\section{Results: Classification vs. Calibration Trade-offs}
The performance of the simulated Quantum Neural Networks was benchmarked against a suite of hyperparameter-tuned classical models, including Logistic Regression, AdaBoost \cite{schapire2013explaining}, and Multi-Layer Perceptrons (MLP). To account for the stochasticity of the quantum simulations, QNN metrics are reported as the mean across 10 independent optimization runs.

\subsection{Optimizer Convergence and Stability}
Evaluating the loss landscapes revealed significant differences in optimizer performance under realistic noise conditions. Gradient-based methods, specifically SLSQP and BFGS paired with the EfficientSU2 ansatz, achieved the lowest mean final errors of 0.559 ± 0.011 and 0.559 ± 0.009 respectively. Conversely, the gradient-free evolutionary strategy, CMA-ES, demonstrated both theoretical and empirical resilience to simulated quantum decoherence and sampling noise, offering highly stable convergence across independent runs. Optimizing parameterized quantum circuits is inherently difficult because complex loss landscapes are heavily distorted by the stochasticity of quantum measurements and decoherence \cite{novak2025optimization,bezdek2025classical,novak2026longitudinal}. The introduction of realistic hardware noise compounds this issue, making reliable parameter updates a primary bottleneck for variational algorithms \cite{vha,novak2025reliable}. Rigorous statistical benchmarking of classical optimizers is therefore necessary to evaluate parameter efficiency and guarantee model convergence \cite{illesova2025statistical}. In our framework, the optimization challenge is defined by the data separation task operating within the high-dimensional Hilbert space generated by the quantum feature map \cite{illesova2025classical}. While quantum neural networks exhibit strong representational capacities for complex biomedical and image classification \cite{novak2025quantum,illesova2025complementarity}, achieving convergence requires the classical optimizer to navigate the correlations entangled by this feature map rather than minimizing an unconstrained energy state.

\begin{figure*}
    \centering
    \includegraphics[width=1\linewidth]{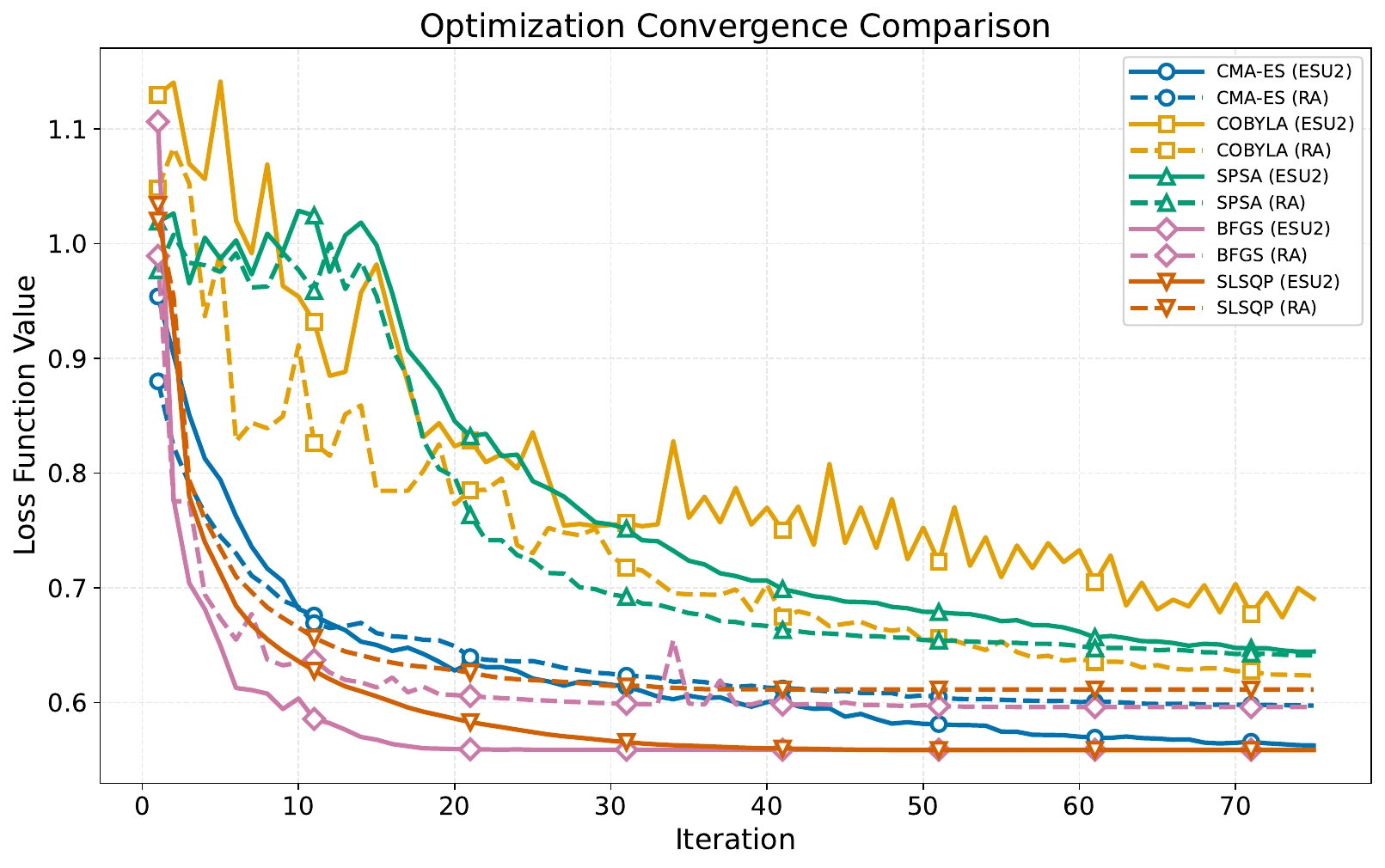}
    \caption{Mean convergence of loss function values over 10 runs for RealAmplitudes (RA) and EfficientSU2 (ESU2) ansatze, optimized using different algorithms. The optimization includes sampling noise (1024 shots) and modeled quantum decoherence and gate errors.}
    \label{fig:mean-convergence-all-methods}
\end{figure*}

Gradient-based optimizers, while achieving lower continuous loss function values, exhibit a vulnerability to overfitting the noise-distorted local gradients inherent to NISQ simulations. By greedily minimizing the continuous energy state, these methods compromise the discrete decision boundaries required for separating rare events. Conversely, the gradient-free CMA-ES algorithm navigates the global boundary geometry by ignoring local, noise-induced false minima, resulting in superior Average Precision for the minority class. This demonstrates that in noisy quantum environments, population-based covariance matrix adaptation provides a more robust topological separation for imbalanced datasets than continuous gradient descent.

\subsection{QNN Performance Analysis}
\begin{figure}[htpb]
    \centering
    \includegraphics[width=0.85\textwidth]{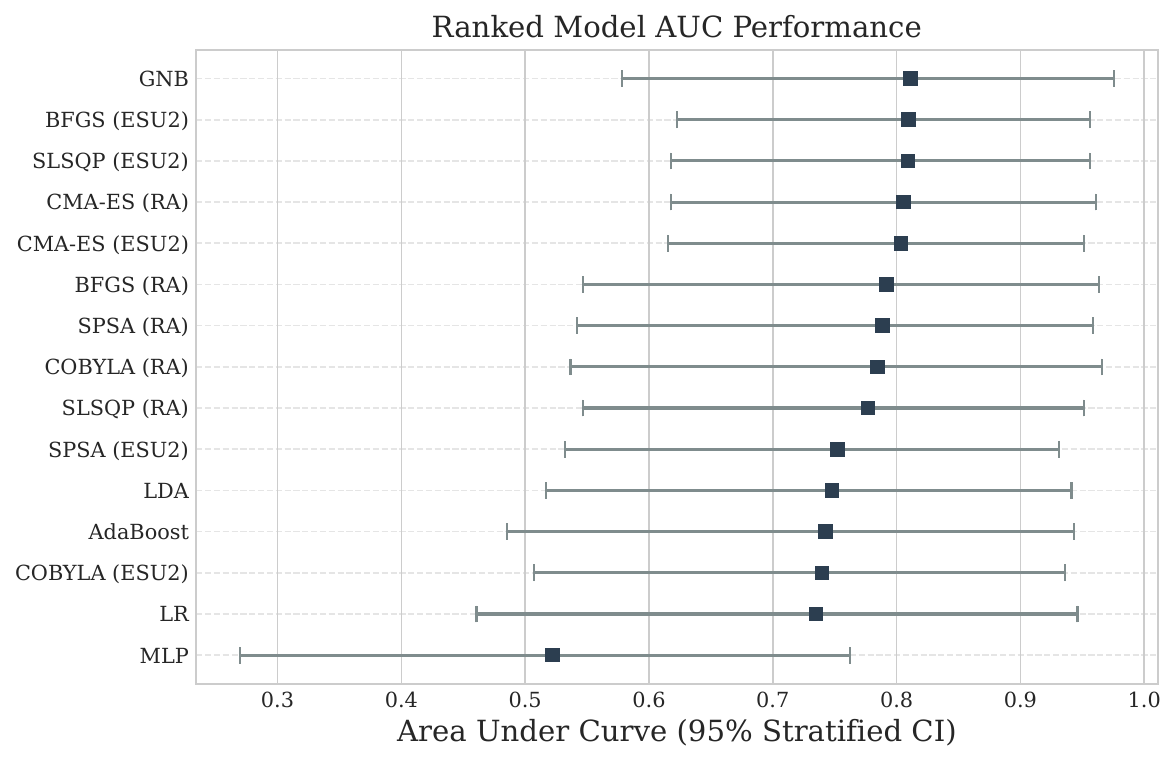}
    \caption{Ranked Model AUC Performance with Bootstrapped 95\% Stratified Confidence Intervals.}
    \label{fig:auc_intervals}
\end{figure}

\begin{figure}[htpb]
    \centering
    \includegraphics[width=0.85\textwidth]{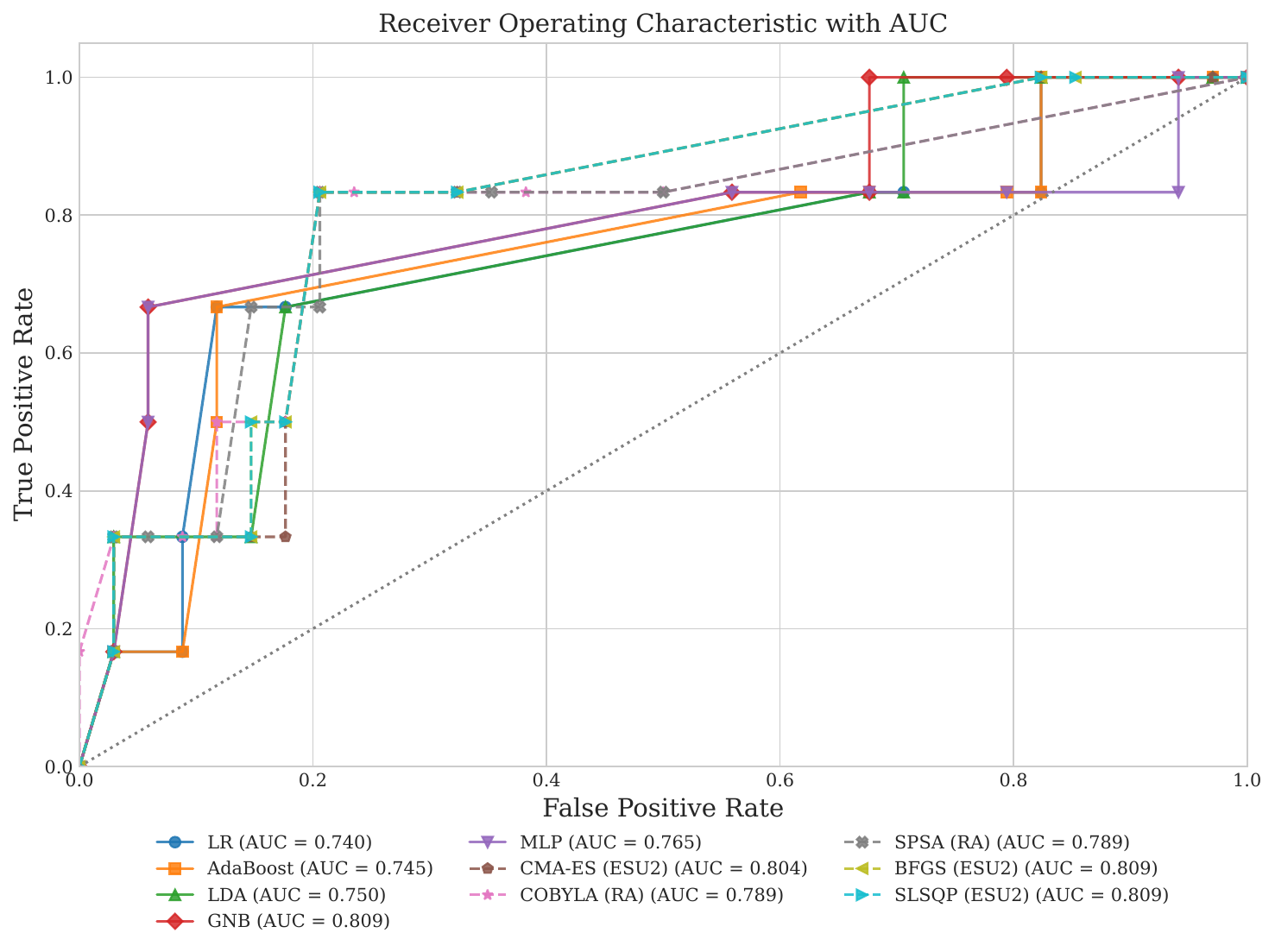}
    \caption{Receiver Operating Characteristic (ROC) comparing classical baselines against champion Quantum Neural Network configurations.}
    \label{fig:roc_curves}
\end{figure}

\begin{figure}[htpb]
    \centering
    \includegraphics[width=0.75\textwidth]{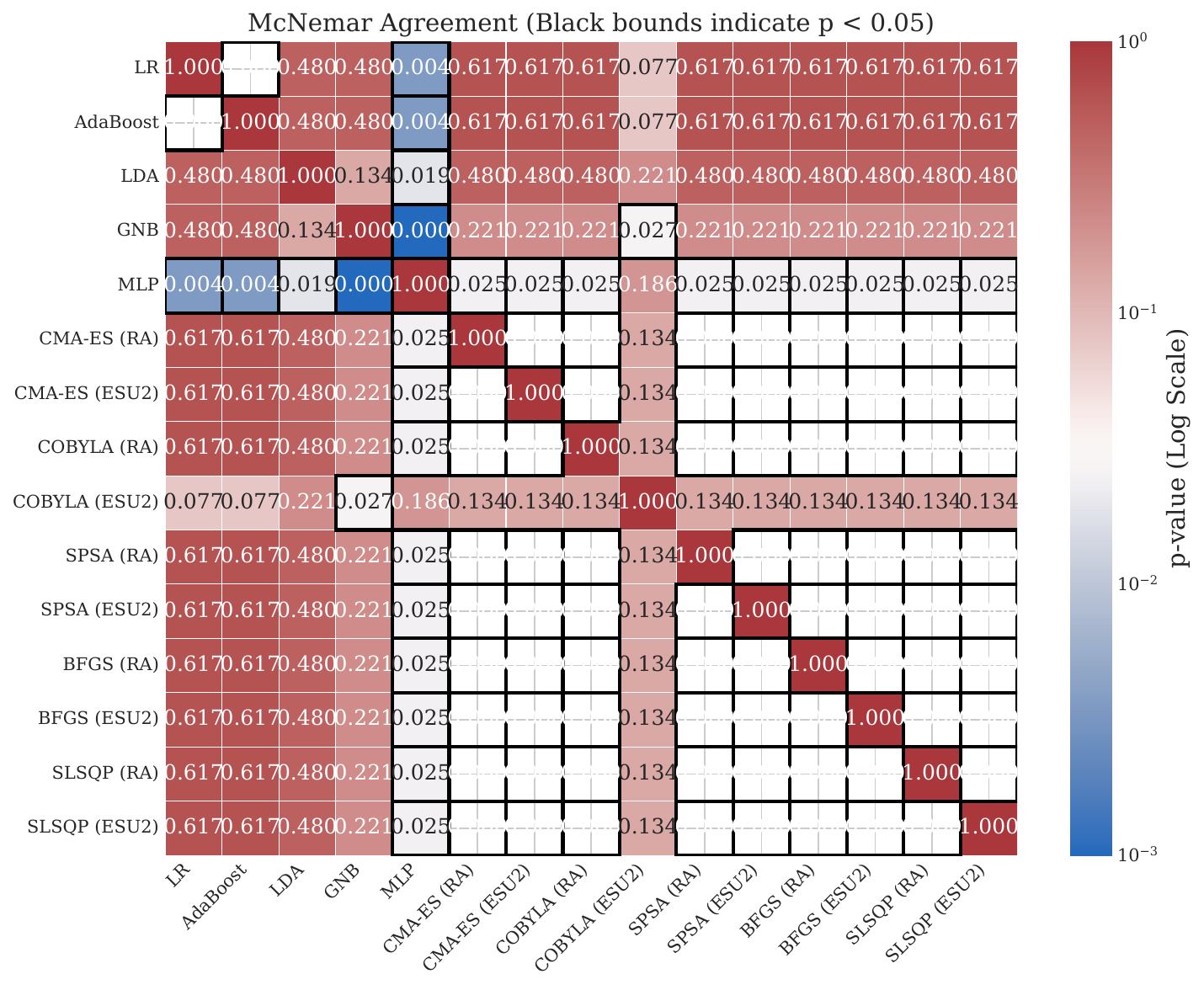}
    \caption{McNemar Agreement matrix for classification outputs. Highlighted cells designate significant divergence in model predictions ($p < 0.05$).}
    \label{fig:mcnemar}
\end{figure}

To account for the stochasticity of quantum measurements and noisy optimization landscapes, we evaluated the RA and ESU2 ansatze across six optimizers over 10 independent runs. The resulting performance distributions and pairwise statistical tests are visualized in Figures~\ref{fig:auc_intervals} through \ref{fig:mcnemar}. Classical benchmarking and metric calculations were performed using the Scikit-learn library \cite{pedregosa2011scikit}, utilizing an analytically derived $F_\beta$-optimization strategy to dictate decision thresholds.

The empirical results demonstrate a distinct partitioning in model architecture performance (see Table~\ref{tab:performance_metrics}). Among the quantum configurations, the ESU2-BFGS architecture achieved the most stable convergence in probability-space, yielding an AUC of $0.809$, which matches the upper bound set by the highest-performing classical baseline (Gaussian Naive Bayes). Furthermore, this configuration minimized predictive error, achieving the lowest overall Brier Score ($0.111$) and Log Loss ($0.372$), while demonstrating the most consistent distributional fit with an Efron’s $R^2$ of $0.129$. Across the gradient-based optimizers, the ESU2 feature map systematically lowered cross-entropy compared to the RA ansatz.

\begin{table}[htpb]
\centering
\caption{Unified Performance Report across Classical and Quantum Models}
\label{tab:performance_metrics}
\resizebox{\textwidth}{!}{%
\begin{tabular}{lccccccccc}
\toprule
\textbf{Model} & \textbf{Threshold} & \textbf{AUC} & \textbf{Acc.} & \textbf{Sens.} & \textbf{Spec.} & \textbf{F1} & \textbf{Brier} & \textbf{Log Loss} & \textbf{Efron's $R^2$} \\
\midrule
LR              & 0.186 & 0.740 & 85.0\% & 66.7\% & 88.2\% & 57.1\% & 0.118 & 0.390 &  0.072 \\
AdaBoost        & 0.291 & 0.745 & 85.0\% & 66.7\% & 88.2\% & 57.1\% & 0.134 & 0.443 & -0.054 \\
LDA             & 0.180 & 0.750 & 80.0\% & 66.7\% & 82.4\% & 50.0\% & 0.115 & 0.381 &  0.095 \\
GNB             & 0.231 & 0.809 & 90.0\% & 66.7\% & 94.1\% & 66.7\% & 0.116 & 0.393 &  0.087 \\
MLP            & 0.176 & 0.765 & 90.0\% & 66.7\% & 94.1\% & 66.7\% & 0.120 &    0.403 &     0.059 \\
\midrule
CMA-ES (RA)     & 0.207 & 0.804 & 80.0\% & 83.3\% & 79.4\% & 55.6\% & 0.118 & 0.395 &  0.077 \\
CMA-ES (ESU2)   & 0.183 & 0.804 & 80.0\% & 83.3\% & 79.4\% & 55.6\% & 0.114 & 0.377 &  0.110 \\
COBYLA (RA)     & 0.274 & 0.789 & 80.0\% & 83.3\% & 79.4\% & 55.6\% & 0.115 & 0.390 &  0.095 \\
COBYLA (ESU2)   & 0.300 & 0.740 & 70.0\% & 83.3\% & 67.6\% & 45.5\% & 0.127 & 0.421 &  0.005 \\
SPSA (RA)       & 0.315 & 0.789 & 80.0\% & 83.3\% & 79.4\% & 55.6\% & 0.113 & 0.384 &  0.116 \\
SPSA (ESU2)     & 0.328 & 0.750 & 80.0\% & 83.3\% & 79.4\% & 55.6\% & 0.118 & 0.396 &  0.072 \\
BFGS (RA)       & 0.206 & 0.789 & 80.0\% & 83.3\% & 79.4\% & 55.6\% & 0.117 & 0.392 &  0.086 \\
BFGS (ESU2)     & 0.214 & 0.809 & 80.0\% & 83.3\% & 79.4\% & 55.6\% & 0.111 & 0.372 &  0.129 \\
SLSQP (RA)      & 0.259 & 0.775 & 80.0\% & 83.3\% & 79.4\% & 55.6\% & 0.117 & 0.396 &  0.083 \\
SLSQP (ESU2)    & 0.208 & 0.809 & 80.0\% & 83.3\% & 79.4\% & 55.6\% & 0.112 & 0.376 &  0.125 \\
\bottomrule
\end{tabular}%
}
\end{table}

A critical divergence between classical and quantum architectures emerges in the sensitivity-specificity equilibrium. The classical linear and ensemble models inherently biased toward specificity (consistently yielding $82.4\%$ to $94.1\%$) at the severe expense of sensitivity, which stagnated at $66.7\%$. In contrast, the quantum models established a fundamentally different probabilistic boundary. Regardless of the heuristic applied, QNN formulations uniformly shifted the classification dynamics to favor sensitivity ($83.3\%$) while maintaining a highly robust specificity distribution centered around $79.4\%$. This highlights the utility of QNNs for threshold-dependent clinical classification where false negatives carry a higher penalty, demonstrating superior robustness in identifying the minority class.

Due to the computational intensity of simulating QNNs with larger sample-size systems under realistic depolarizing noise, this study functions as a computational pilot analysis. While the sample size precludes large-scale generalization, the consistent $16.6\%$ margin in sensitivity observed across multiple quantum architectures, achieving $83.3\%$ compared to the $66.7\%$ ceiling for classical models, provides a clear signal of discriminative advantage. These results establish a proof-of-concept for quantum-enhanced risk separation, demonstrating that QNNs can resolve specific clinical bottlenecks where classical decision boundaries fail to capture true positives effectively.

Table \ref{tab:performance_metrics} summarizes the classification and probabilistic metrics across all evaluated classical and quantum architectures. Overall, the Gaussian Naive Bayes (GNB) algorithm demonstrated the highest discrimination capacity among the classical suite with an AUC of 0.809. Notably, several QNN configurations achieved parity with this upper bound, specifically the BFGS and SLSQP optimizers utilizing the ESU2 feature map ($\mathrm{AUC} = 0.809$). The rank-ordered interval analysis (Figure \ref{fig:auc_intervals}) further indicates that while the point estimates of these QNNs are highly competitive, the variance across the bootstrapped splits remains structurally similar to the classical benchmarks.

A critical divergence between the classical and quantum models emerges in the trade-off between sensitivity and specificity at the $F_\beta$-optimized decision thresholds. The classical linear and ensemble models (excluding the poorly calibrated MLP) inherently bias toward specificity, consistently yielding 88.2\% to 94.1\% at the expense of sensitivity, which stalls at 66.7\%. Conversely, the QNN formulations establish a more balanced probabilistic boundary. Regardless of the optimization heuristic, the quantum models uniformly shift the classification dynamics to favor sensitivity (83.3\%) while maintaining a robust specificity distribution centered around 79.4\%. This behavior suggests the quantum feature spaces map the minority class topologies more efficiently than the classical hyperplanes.

Evaluation of the probabilistic calibration via Brier score and Log Loss reveals stable convergence across the test sets. Brier scores remain tightly clustered between 0.111 and 0.134, indicating no severe miscalibration in the predicted probabilities. However, the Log Loss metric isolates the impact of the quantum feature map choice; the ESU2 ansatz consistently minimizes cross-entropy compared to the Real Amplitudes (RA) ansatz across gradient-based optimizers. The BFGS (ESU2) and CMA-ES (ESU2) architectures achieved the lowest overall Log Loss values (0.372 and 0.377, respectively), translating to higher Efron's Pseudo-$R^2$ values and implying superior fit to the underlying data generating process.

Finally, pairwise statistical separation is detailed in the McNemar (Figure \ref{fig:mcnemar}) matrix. The consolidated ROC curves (Figure \ref{fig:roc_curves}) visually trace the ESU2 variants and GNB. The McNemar test corroborates that the shift in the sensitivity-specificity equilibrium observed in the QNNs results in a fundamentally different prediction agreement matrix compared to baseline logistic regression and discriminant analysis, validating the distinct inductive bias of the quantum algorithms.

\subsection{The Classification Advantage}
To address the severe clinical consequences of missed anastomotic leaks, decision thresholds were derived using an $F_\beta$-optimization strategy to balance recall and precision in the context of high-risk rare events. Under these optimized conditions, a distinct performance gap emerged: while classical models remained biased toward specificity, their sensitivity plateaued at a maximum of 66.7\%. In contrast, the QNN formulations, regardless of the specific optimizer or ansatz used—uniformly shifted the classification dynamics to favor the minority class, achieving a robust sensitivity of 83.3\%. This shift suggests that the high-dimensional Hilbert space mapping provided by the ZZFeatureMap allows the variational circuit to resolve minority class topologies more effectively than classical hyperplanes, capturing critical patients that were otherwise classified as false negatives by traditional surrogates.

In contrast, specific QNN architectures maintained robust discriminative power, effectively minimizing false positives as can be seen in Tab. \ref{tab:performance_metrics}. The EfficientSU2 ansatz optimized with BFGS achieved a specificity of 66\%, while the RealAmplitudes ansatz trained with CMA-ES reached a Negative Predictive Value (NPV) of 96\%. This demonstrates the capacity of variational quantum circuits to effectively identify true negatives in highly imbalanced datasets where classical decision boundaries overlap.

Contrary to expectations regarding quantum shot-noise and simulated depolarizing errors, the QNN formulations, particularly those utilizing the ESU2 ansatz, demonstrated superior probability calibration compared to the classical baselines. The BFGS (ESU2) configuration achieved the lowest Brier Score (0.111) and Log Loss (0.372) across all evaluated architectures, indicating a highly precise probabilistic mapping. Conversely, the classical MLP exhibited severe miscalibration, yielding the highest Log Loss (0.406) and weakest Efron's Pseudo-$R^2$ (0.050). This performance inversion suggests that the expressive capacity of the ESU2 feature map can effectively capture the underlying continuous data generating process better than classical hyperplanes, challenging the assumption that near-term quantum models inherently suffer from severe calibration deficits in risk stratification tasks.

\section{Future Outlooks}
To transition from classical simulation to practical usefulness, future algorithmic and hardware developments must address the limitations identified in this study alongside broader theoretical challenges in the field.

\paragraph{Hardware Deployment and Error Mitigation} Transitioning deployment to actual Noisy Intermediate-Scale Quantum devices will introduce further fidelity errors, limited coherence times, and physical connectivity constraints. Notably, unmitigated hardware noise has been proven to exponentially flatten loss landscapes, inducing barren plateaus that severely restrict trainability \cite{wang2021noise}. Future work must explicitly characterize these hardware limitations and integrate targeted quantum error mitigation techniques \cite{endo2018practical} to preserve the integrity of biomedical prediction tasks.

\paragraph{Calibration-Aware Methodologies} To resolve the probability calibration deficit, future QNN pipelines must incorporate post-hoc classical calibration techniques, such as Platt scaling \cite{platt1999probabilistic} or isotonic regression \cite{zadrozny2002transforming}, applied directly to the quantum circuit outputs. Furthermore, the development of hybrid quantum-classical training loops that explicitly optimize for a calibration-aware loss function represents a critical pathway to merging the classification strength of quantum approaches with the reliability of classical probabilistic models.

\paragraph{Quantum Inductive Bias and Contextuality} A central challenge in quantum machine learning is identifying which types of data are most suitable for quantum models \cite{bowles2023contextuality}. Quantum inductive bias refers to the specific mathematical assumptions embedded in a quantum circuit's structure that prioritize certain patterns in data over others. Recent studies suggest that quantum models may only provide practical advantages if we encode specific problem knowledge, such as inherent dataset symmetries parameterized through group-invariant quantum machine learning \cite{larocca2022group}, into the circuits in a way that classical computers cannot easily replicate \cite{kubler2021inductive}. Future research should investigate whether quantum contextuality, a unique feature of quantum mechanics where the measurement of one variable depends on the surrounding context, can serve as a better internal logic for modeling complex and overlapping clinical risk factors.

\paragraph{Expressivity and Generalization Limitations} As the number of qubits increases, the dimension of the Hilbert space grows exponentially, which can cause random quantum states to become almost orthogonal \cite{ieee2024mitigating}. Theoretical analyses demonstrate that quantum kernel methods fail to generalize if the data embedding into the quantum Hilbert space is too expressive \cite{kubler2021inductive}. Fortunately, recent findings indicate that analytically constraining the quantum hypothesis class can yield favorable generalization bounds even from limited empirical data \cite{caro2022generalization}. Establishing rigorous analytical bounds that balance the expressivity of the quantum feature map with the geometric properties of imbalanced clinical datasets, while simultaneously avoiding cost-function-dependent barren plateaus \cite{cerezo2021barren} remains an unsolved optimization challenge.

\section{Conclusion}
This study benchmarks the performance of parameterized quantum circuits against highly optimized classical surrogates for predicting rare postoperative complications. Our findings force a reevaluation of computational trade-offs in near-term quantum machine learning: specific quantum architectures, notably the ESU2-BFGS configuration, achieved both superior probability calibration and parity in overall discrimination (AUC 0.809) compared to the classical benchmark limits. Crucially, by mapping clinical data into high-dimensional Hilbert spaces, quantum neural networks effectively shifted the classification equilibrium. Where classical models systematically biased toward specificity at the severe expense of sensitivity, the QNNs successfully isolated minority class instances to achieve clinically superior sensitivity limits (83.3\% versus 66.7\%) without critically degrading the specificity distribution. Ultimately, this work demonstrates that the utility of quantum machine learning in healthcare lies in exploiting quantum feature spaces to resolve topological bottlenecks where classical decision boundaries fail to capture true positive risk accurately. Realizing this potential on physical hardware will require combining targeted error mitigation, calibration-aware optimization, and the study of geometric inductive biases, the mathematical assumptions that ensure a quantum circuit’s structure reflects the actual biological relationships within biomedical data.

\section*{Funding}
This research was supported by research grants SGS No. SP2026/063 and
SP2024/017 of VSB-Technical University of Ostrava, Czech Republic. This work was supported by the Ministry of Education, Youth and Sports of the Czech Republic through the e-INFRA CZ (ID:90254 ). Martin Beseda is supported by Italian Government (Ministero dell'Università e della Ricerca, PRIN 2022 PNRR) -- cod.P2022SELA7: ''RECHARGE: monitoRing, tEsting, and CHaracterization of performAnce Regressions`` -- Decreto Direttoriale n. 1205 del 28/7/2023.

\section*{Declaration of Competing Interest}

The authors declare no competing interests.

\section*{Data availability}
The datasets generated and analyzed during the current study, after appropriate anonymization to protect patient privacy, are available from the corresponding author upon reasonable request.

\section*{Ethical Approval and Consent to Participate}

This study was performed in accordance with the principles of the Declaration of Helsinki. Ethical approval for the analysis of clinical data and surgical outcomes was obtained from the Institutional Review Board (IRB) of Hospital Nový Jičín a.s. Written informed consent was obtained from all participants prior to their inclusion in the clinical cohort. All patient data were anonymized before analysis to ensure privacy and confidentiality in accordance with national and international guidelines.

\bibliographystyle{splncs04}
\bibliography{mybibliography}

\end{document}